%% file: main.tex
\documentclass[11pt]{article}
\usepackage[margin=1in]{geometry}
\usepackage{graphicx}
\usepackage{booktabs}
\usepackage{enumitem}
\usepackage{amsmath}
\usepackage{amssymb}
\usepackage{algorithm}
\usepackage{algpseudocode}
\usepackage{listings}
\usepackage{xcolor}
\usepackage{hyperref}
\usepackage{caption}
\usepackage{subcaption}
\usepackage{multirow}
\usepackage{rotating}
\usepackage{array}
\usepackage{microtype}
\usepackage{seqsplit}
\usepackage{xspace}
\usepackage{tikz}
\usepackage{pgfplots}
\usetikzlibrary{patterns,arrows.meta,positioning}
\usepgfplotslibrary{groupplots}
\pgfplotsset{compat=1.18}
\definecolor{codebg}{RGB}{248,248,248}
\definecolor{codeborder}{RGB}{204,204,204}
\definecolor{rlink}{RGB}{33,40,48}
\definecolor{rlaxis}{RGB}{88,96,105}
\definecolor{rlgrid}{RGB}{224,229,235}
\definecolor{rlteal}{RGB}{0,121,130}
\definecolor{rlblue}{RGB}{43,101,151}
\definecolor{rlslate}{RGB}{116,128,140}
\DeclareRobustCommand{\system}{\textnormal{\textsc{ResearchLoop}}\xspace}

\lstdefinelanguage{yaml}{
  keywords={true,false,null,y,n},
  keywordstyle=\color{blue}\bfseries,
  basicstyle=\ttfamily\small,
  morecomment=[l]{\#},
  commentstyle=\color{gray}\itshape,
  morestring=[b]',
  morestring=[b]",
  stringstyle=\color{red},
  literate={:}{:}1,
  breaklines=true,
  showstringspaces=false,
  tabsize=2
}
\title{\system: An Evidence-Gated Control Plane for AI-Assisted Research\\
{\large Technical Report}}
\author{
Yihan Xia\\
Shenzhen University, China\\
\texttt{xiayihan2023@email.szu.edu.cn}
\and
Taotao Wang\\
Shenzhen University, China\\
\texttt{ttwang@szu.edu.cn}
}
\date{May 2026}

\begin{document}
\maketitle

\begin{abstract}
AI-assisted research compresses ideation, implementation, evaluation, and manuscript writing into a single interactive loop. This compression is useful, but it also creates a publication risk: paper claims can become easier to state than to audit. We present \emph{\system}, an evidence-gated control plane for AI-assisted computational research. \system treats research questions, task contracts, evidence objects, claim ledgers, closeouts, and paper bindings as durable project state, realized here as a repository-backed runtime. This technical report provides the complete protocol specification, state model, transition rules, claim-admission algorithm, and insight-compounding mechanism. It also reports the full experimental record spanning nine versions (V0--V9), including a self-hosting case study, a controlled task-suite study with component ablations, a mathematical olympiad evaluation, and a supplementary SciCode boundary experiment evaluated with the official generated-code harness. All artifacts, manifests, and verification reports are preserved in the project repository.
\end{abstract}

\tableofcontents
\newpage

\input{introduction}
\input{related_work}
\input{design}
\input{runtime}
\input{evidence_gates}
\input{insight_compounding}
\input{evaluation}
\input{threats}
\input{conclusion}
\input{artifact_availability}

\appendix
\input{protocol_spec}
\input{experiment_details}
\input{claim_ledger_full}

\bibliographystyle{plain}
\bibliography{references}

\end{document}

%% file: introduction.tex
\section{Introduction}
\label{sec:introduction}

AI-assisted research workflows compress ideation, implementation, evaluation, and manuscript writing into a single interactive loop. A single researcher can ask language agents to inspect repositories, write code, run commands, summarize evidence, draft related work, and propose paper narratives within the same session. This compression is operationally attractive, but it creates a software-engineering control problem: the manuscript can become the most complete-looking artifact before the evidence base is ready to support it.

The failure mode is not merely hallucination. It is a \emph{state-management failure} in which execution artifacts, paper claims, and research questions drift out of synchronization. An agent can perform locally plausible work that no longer answers the active research question. A command can run and a table can be generated without supporting the claim the manuscript wants to make. Language models can stabilize a coherent paper story before the research state is mature. Failed reproductions, blocked baselines, and rejected claims can disappear from the narrative rather than becoming typed state for the next version.

These failures suggest a compact design requirement for AI-assisted computational research: every paper-facing claim should be traceable to the research question that authorized the work, the evidence object that supports it, the gate that admitted it, and the version boundary that preserved its scope.

\emph{\system} makes this state explicit. It is a control plane---not a stronger agent, a longer context window, or a better prompt template---that separates executors from the persistent rules that decide what is runnable, what counts as evidence, what may become a paper claim, and what must be carried forward after a version closes. \system is organized around three design commitments. First, it is \emph{RQ-driven}: tasks are authorized by research questions rather than by an unstructured to-do list. Second, it is \emph{evidence-gated}: claims do not enter result language merely because artifacts exist; they must pass an explicit evidence contract and admission decision. Third, it is \emph{insight-compounding}: failures, anomalies, blocked baselines, and rejected claims become durable state for future versions rather than disappearing from the narrative.

This technical report is the complete account of the \system protocol and its empirical record. It contains the full specification, the implementation runtime, the self-hosting case study, and all controlled experiments from V0 through V9. Unlike a conference submission, which must compress nine versions into bounded claims, this report preserves the complete research trajectory including negative results, blocked claims, and version pivots.

\subsection{Contributions}

This report makes four contributions.

\begin{enumerate}[leftmargin=*]
    \item \textbf{Control-plane abstraction.} It formulates long-horizon AI-assisted research as a control-plane problem in which RQs, execution contracts, evidence objects, insights, and paper claims must remain synchronized across version boundaries.
    \item \textbf{Protocol specification.} It specifies \system as a state-transition protocol with durable objects, gate-bounded claim admission, blocker preservation, and closeout-driven reseeding.
    \item \textbf{Conservative publication interface.} It defines a publication interface in which design, process, controlled-study, and supplementary boundary claims are admitted only at the scope licensed by their evidence gates.
    \item \textbf{Complete empirical record.} It reports the full V0--V9 experimental trajectory, including the self-hosting case study, the controlled task-suite study with component ablations, the mathematical olympiad evaluation (V4), the SciCode boundary experiments (V8--V9), and all closeout records with blocker preservation.
\end{enumerate}

%% file: related_work.tex
\section{Related Work}
\label{sec:related-work}

\system sits at the intersection of autonomous research agents, structured research workflows, experiment and provenance infrastructure, paper-writing systems, research-agent benchmarks, and empirical software-engineering methodology. The contribution is not that planning, logging, review, or manuscript generation are new. The contribution is a publication-facing control plane in which RQs schedule work, evidence gates regulate claim admission, closeouts preserve version boundaries, and paper bindings project only admitted claims into manuscript text.

\paragraph{End-to-end AI scientists.}
Recent systems increasingly make large portions of the scientific workflow executable. The AI Scientist and AI Scientist-v2 automate ideation, coding, experimentation, paper writing, and review for machine-learning research~\cite{lu2024aiscientist,yamada2025aiscientistv2}. Agent Laboratory, ARIS, and AlphaEvolve similarly explore broader agentic research loops, multi-agent collaboration, or algorithmic discovery~\cite{schmidgall2025agentlaboratory,yang2026aris,novikov2025alphaevolve}. In science-facing settings, Co-Scientist focuses on structured hypothesis generation and biomedical validation, Robin combines literature-search and data-analysis agents for experimental biology, and ERA uses LLM-guided tree search to write and optimize empirical scientific software~\cite{gottweis2026coscientist,ghareeb2026robin,aygun2026era}.

These systems show that agents can propose ideas, execute code, analyze results, and draft scientific narratives. \system addresses a different control problem. Its primary object is not the autonomous worker but the paper-facing state transition from candidate claim to admitted claim. A successful run, a generated figure, or a plausible review round remains insufficient unless the claim's evidence contract and gate predicate are satisfied at the declared scope.

\paragraph{Structured research frameworks and gate-based pipelines.}
The closest systems to \system are structured research frameworks that make stages, gates, state, or memory explicit. PARNESS provides a paper harness with dynamic workflows, full-text indexing, and cross-run knowledge accumulation~\cite{wang2026parness}. Sibyl-AutoResearch argues that autonomous research needs self-evolving trial-and-error harnesses; its file-backed SIBYL system records state, roles, memory, gates, artifact traces, and recovered failures~\cite{wang2026sibyl}. These systems are especially relevant because they move beyond single-shot paper generation toward persistent research state and reusable trial experience.

\system differs in where it places the gate. In many stage-based frameworks, a gate advances a workflow: a plan becomes runnable, a run becomes analyzable, or a draft becomes reviewable. In \system, the privileged transition is manuscript-facing: a claim becomes paper-bindable only if its RQ link, evidence object, gate decision, scope, and version boundary are recorded. This distinction matters because research automation can be operationally successful while still over-admitting claims. \system therefore treats blocked, contradicted, and forbidden claims as durable states, not merely failed workflow steps.

\paragraph{Paper-writing and evidence-to-manuscript systems.}
Automated writing systems address the final compression from research materials to paper text. PaperOrchestra transforms unconstrained pre-writing materials into submission-ready LaTeX manuscripts and evaluates writing quality through PaperWritingBench~\cite{song2026paperorchestra}. Data-to-paper starts from annotated data and produces transparent, backward-traceable, human-verifiable scientific papers with data-chained links from manuscript values to code and data~\cite{ifargan2024datatopaper}. These systems are important because they make manuscript generation more evidence-aware than ordinary free-form prompting.

\system is complementary to such writing engines. A writing engine can assemble sections, citations, tables, and narratives; \system decides which claims are licensed for assertive result language. Data-to-paper's data chaining is strongest for numerical traceability from data and code to manuscript values. \system generalizes the publication boundary to heterogeneous claim classes: design claims, process claims, audit claims, empirical effectiveness claims, and blocked future claims each require different evidence contracts.

\paragraph{Experiment automation, provenance, and workflow systems.}
Experiment and provenance infrastructure supplies the artifacts that \system consumes. PROV-DM models entities, activities, and agents in artifact production~\cite{moreau2013prov}. MLflow and DVC organize machine-learning runs, datasets, metrics, and artifacts~\cite{zaharia2018mlflow,kuprieiev2021dvc}. Nextflow and Snakemake support reproducible computational workflows~\cite{di2017nextflow,koster2018snakemake}. Git-backed task agents such as SWE-agent, OpenHands, Devin, and Aider operate in the same repository-centered execution niche as \system and produce durable diffs, test outcomes, and workspace traces~\cite{yang2024sweagent,wang2024openhands,cognition2024devin,gauthier2023aider}.

These tools provide lineage, execution machinery, or task-level durability, but they do not by themselves answer the publication question: which manuscript claim does this artifact support, under what scope, and after which gate decision? \system treats their outputs as evidence objects rather than as automatic claim licenses. A completed workflow, a passing test suite, or a Git commit can support a claim only relative to a claim class and evidence contract.

\paragraph{Claim-evidence benchmarks.}
Research-agent benchmarks and claim-evidence benchmarks motivate finer-grained evaluation. ResearchGym and AIRS-Bench define bounded research tasks for AI research agents~\cite{garikaparthi2026researchgym,lupidi2026airs}. FIRE-Bench, DeepFact, and CLAIM-BENCH emphasize factuality, rediscovery, and scientific claim--evidence reasoning~\cite{wang2026firebench,huang2026deepfact,javaji2025claimbench}. \system borrows the need for task- and claim-level measurement, but its immediate contribution is not a new benchmark. It specifies when benchmark outcomes, audit labels, or run artifacts may become paper-admissible claims.

\paragraph{Agent memory and operating-system analogies.}
A related architectural line treats LLM agents as operating systems or resource-managed agents. MemGPT introduces hierarchical memory management for long-horizon agent execution, and AgentRM proposes an OS-inspired resource manager for agent systems~\cite{packer2024memgpt,she2026agentrm}. These systems manage object-level agent resources such as context, memory, and compute. \system is a meta-level control plane: it manages research-process state such as RQs, evidence, blockers, claim ledgers, closeouts, and paper bindings. The two abstractions are complementary; better agent memory can improve execution, while \system regulates which executions may become manuscript claims.

\paragraph{Empirical SE method.}
Evidence-based software engineering, artifact-evaluation practice, and case-study methodology stress proportionate claims, validity threats, reproducibility, and inspectable artifacts~\cite{kitchenham2004evidence,treude2026artifactevaluation,runeson2009guidelines}. \system moves this discipline into the research loop rather than leaving it as an end-of-paper checklist. Its claim ledger and evidence gates operationalize a conservative writing rule: evidence first, admitted scope second, paper language last.

\begin{table}[t]
\centering
\scriptsize
\begin{tabular}{@{}p{0.34\linewidth}ccccc@{}}
\toprule
\textbf{System family} & \textbf{RQ} & \textbf{Evidence} & \textbf{Ledger} & \textbf{Gate} & \textbf{Paper} \\
\midrule
End-to-end AI scientists & P & P & P & P & P \\
Structured research frameworks & P & Y & P & P & P \\
Paper-writing systems & P & P & P & P & Y \\
Experiment/provenance tools & -- & Y & -- & P & -- \\
Git-backed task agents & P & Y & -- & P & -- \\
Benchmarks and claim-evidence tasks & C & P & C & C & -- \\
Empirical-method guidance & C & C & C & C & C \\
\system & Y & Y & Y & Y & Y \\
\bottomrule
\end{tabular}
\caption{Positioning of \system relative to adjacent system families. Y denotes explicit protocol support, P partial or indirect support, C conceptual guidance, and -- no primary support. The table is intentionally coarse; the prose explains the ratings.}
\label{tab:related-positioning}
\end{table}

\paragraph{Positioning.}
\system should be read as a control architecture for research reliability. Agent systems supply execution, structured frameworks supply workflow discipline, provenance tools supply trace records, writing systems supply manuscript generation, benchmarks supply evaluation pressure, and empirical-method traditions supply validity discipline. \system binds these pieces to RQs and paper claims so that planned, admitted, blocked, contradicted, and forbidden statements remain distinct, inspectable states.

%% file: design.tex
\section{\system Protocol and Runtime}
\label{sec:technical-core}

\system is a control plane for AI-assisted research. It does not replace experiment engines, notebooks, version control, or agent frameworks. Instead, it defines the persistent research state---implemented here as a repository-backed runtime---that determines which work is authorized, which artifacts count as evidence, which claims may be written assertively, and which unresolved observations must survive into the next version.

The ``OS'' metaphor is structural, not functional. \system does not manage hardware, schedule processes, or allocate memory. It borrows three operating-system concepts: (1) \emph{separation of mechanism from policy}---executors (agents, scripts, humans) are mechanism, while gates and ledgers are policy; (2) \emph{durable state beyond any single process}---the evidence base, claim ledger, and closeout records survive individual task executions; (3) \emph{mediated access to privileged transitions}---workers cannot directly promote a claim to admitted status; they must satisfy a gate predicate.

\paragraph{Process-theoretic grounding.}
\system's abstraction can be read as an operationalization of software-process theory rather than control theory. Parnas and Clements argue that a rational design process requires documented design rationale, traceable decisions, and inspectable checkpoints that make the actual process appear systematic even when real development is iterative and opportunistic~\cite{parnas1986rational}. \system treats the closeout as exactly such a checkpoint: an end-of-version record that captures what was decided, on what evidence, and what remains unresolved. The claim ledger serves as durable design rationale, the RQ spine as the requirements traceability structure, and the gate predicates as the admission criteria that Parnas and Clements prescribe for documenting why a design choice is accepted.

\paragraph{Why the control plane must be external.}
A stronger agent or a longer context window could, in principle, maintain an internal checklist of claims and evidence. However, an internal representation is subject to three risks that an external control plane avoids. First, \emph{narrative control}: the same agent that generates a claim can rewrite the ledger that records it, because both live in the same conversation context. Second, \emph{ephemeral state}: when the conversation ends or the context window resets, the internal record may be lost or condensed into a summary that drops blocked claims. Third, \emph{lack of versioned provenance}: an internal checklist cannot produce an auditable history of who changed a claim status, when, and on what evidence. \system therefore externalizes the ledger, gates, and closeout records to the repository, where they survive any single agent session and can be inspected by humans or automated audits without agent mediation.

\subsection{Core Objects}
\label{subsec:core-objects}

\system uses a small set of durable objects. A \emph{research direction} fixes the Big RQ, falsification boundary, and autonomy limits. A \emph{version} $V_n$ is a bounded research epoch with local stop conditions. An \emph{RQ spine} decomposes the direction into sub-RQs. A \emph{task contract} binds a worker action to an RQ, allowed files, expected artifacts, and forbidden claims. An \emph{evidence object} is a durable artifact such as a run report, log, manifest, hash, source dossier, audit label, or closeout note. A \emph{claim ledger} maps candidate claims to statuses such as planned, admitted, blocked, contradicted, or forbidden. A \emph{paper binding} is the final projection from admitted claims into manuscript text.

\begin{table}[htbp]
\centering
\small
\begin{tabular}{@{}p{0.23\linewidth}p{0.33\linewidth}p{0.32\linewidth}@{}}
\toprule
\textbf{Object} & \textbf{Role} & \textbf{Transition rule} \\
\midrule
Version $V_n$ & Bounded research epoch. & Opens from approved direction state; closes before seeding $V_{n+1}$. \\
RQ spine & Scientific scheduling source. & Tasks without RQ binding are invalid research work. \\
Task contract & Binds worker action to scope, artifacts, and file permissions. & Execution may update state only through the declared contract. \\
Evidence object & Inspectable artifact or decision record. & Supports only claim classes admitted by the relevant gate. \\
Claim ledger & Registry of claim status and paper scope. & Status changes only through recorded evidence and audit. \\
Closeout & End-of-version synthesis of evidence, blockers, and insights. & Seeds the next version; unresolved claims remain visible. \\
\bottomrule
\end{tabular}
\caption{Persistent state objects in \system. Workers are intentionally absent from the truth source: they execute contracts but do not own research state.}
\label{tab:research-os-state}
\end{table}

\subsection{RQ-to-Claim Binding}
\label{subsec:rq-driven-execution}

\begin{figure}[htbp]
\centering
\includegraphics[width=\linewidth]{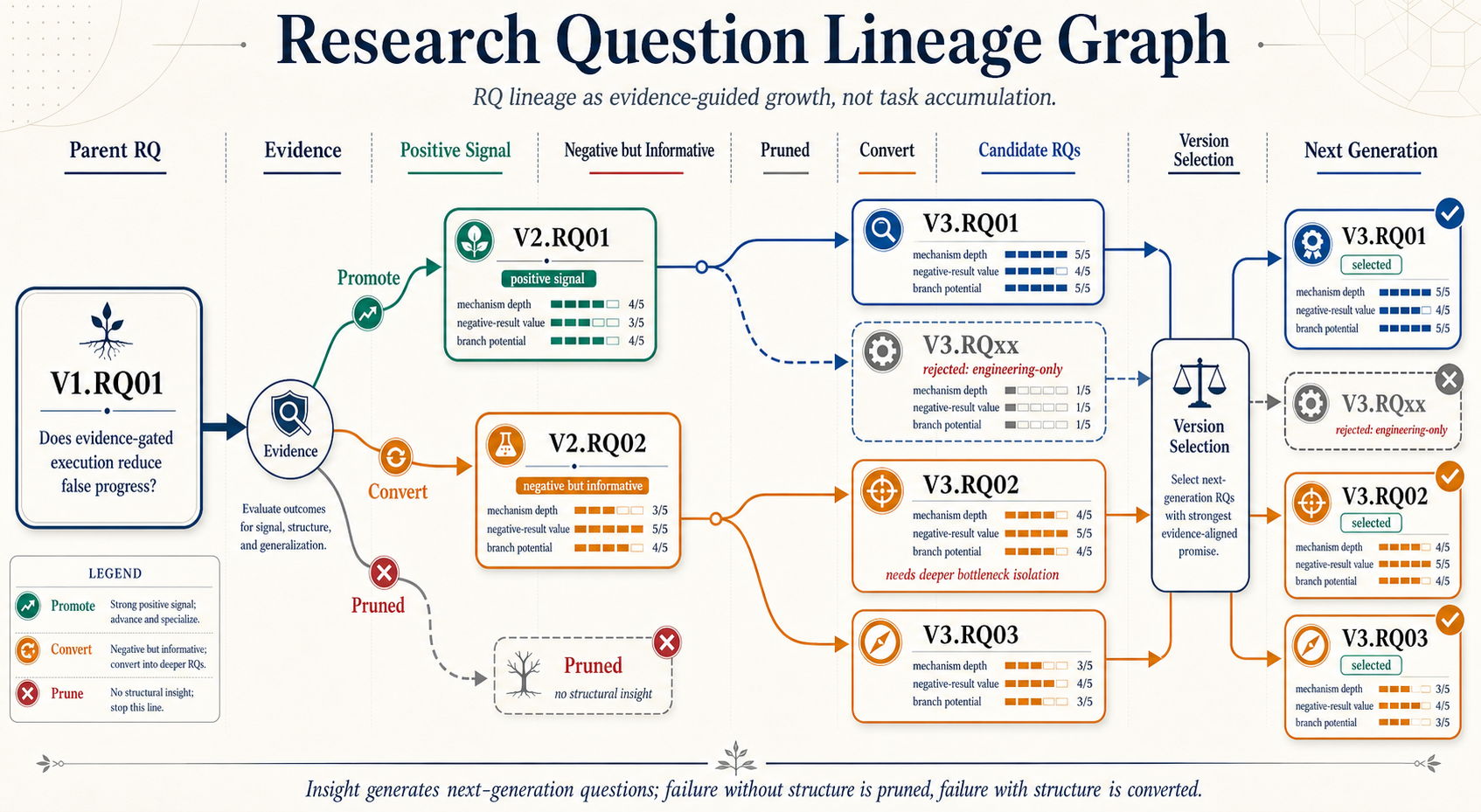}
\caption{Research-question lineage as evidence-guided growth. Positive signals promote a question line, negative-but-informative evidence converts it into deeper questions, and evidence-poor branches are pruned before version selection admits the next generation.}
\label{fig:rq-lineage-graph}
\end{figure}

The central invariant is that paper claims must remain attached to the research question that authorized the work:
\[
\begin{aligned}
\text{Big RQ} &\rightarrow \text{sub-RQ} \rightarrow \text{claim}
\rightarrow \text{task/spec} \\
&\rightarrow \text{evidence} \rightarrow \text{gate}
\rightarrow \text{paper}.
\end{aligned}
\]
The chain is operational, not decorative. Scheduling begins from unresolved RQ state. If a source boundary is missing, the next task is source locking. If evidence exists but lacks audit, the next task is audit. If a claim lacks the required artifact type, the claim remains blocked. The queue is therefore a projection of RQ state rather than an agent-generated to-do list. Figure~\ref{fig:rq-lineage-graph} illustrates how this RQ lineage grows, branches, and prunes before a version admits claims for paper binding.

\subsection{State Model}
\label{subsec:operational-semantics}

Let the state of version $V_n$ be
\begin{equation}
\mathcal{S}_n = \langle \mathcal{D}, \mathcal{R}, \mathcal{K}, \mathcal{E}, \mathcal{L}, \mathcal{B} \rangle,
\end{equation}
where $\mathcal{D}$ is the frozen research direction, $\mathcal{R}$ is the RQ spine, $\mathcal{K}$ is the set of active task contracts, $\mathcal{E}$ is the evidence base, $\mathcal{L}$ is the claim ledger, and $\mathcal{B}$ is the blocker set. A worker execution $w$ under contract $k \in \mathcal{K}$ produces artifacts $A_w$ and report $E_w$. A gate predicate $g(c, A, E_w)$ decides whether evidence $A \subseteq \mathcal{E} \cup A_w$ is sufficient for claim $c$ at its declared scope. In \system, $g$ is a checkable artifact manifest: a declarative list of required artifact types, path patterns, inspection methods, and a pass condition (all or any). The manifest is human-authored or protocol-specified and evaluated deterministically by the runtime.

The transition is:
\begin{equation}
\mathcal{S}_n \xrightarrow{w,k,g} \mathcal{S}'_n = \mathcal{S}_n \oplus \Delta,
\end{equation}
where $\Delta$ may append evidence, update task status, register blockers, and change claim states. It may not rewrite the research direction or promote a claim without a gate decision.

\begin{algorithm}[htbp]
\caption{\system state transition for one task}
\label{alg:researchloop-transition}
\begin{algorithmic}[1]
\Require State $\mathcal{S}_n$, task contract $k$, worker $w$, gate predicate $g$
\Ensure Updated state $\mathcal{S}'_n$
\If{$k$ is not bound to an RQ, claim class, evidence contract, and allowed scope}
    \State reject $k$ as invalid research work
    \State \Return $\mathcal{S}_n$
\EndIf
\State execute $w$ under contract $k$ to produce artifacts $A_w$ and report $E_w$
\State append $A_w$ and $E_w$ to the evidence base as scoped evidence
\If{execution failed or required artifacts are missing}
    \State register blocker $b$; keep affected claims non-admitted
\Else
    \ForAll{claims $c$ linked to $k$}
        \If{$g(c, A_w, E_w)$ is satisfied}
            \State mark $c$ as admitted within the gate-approved scope
        \ElsIf{$g$ identifies contradiction}
            \State mark $c$ as contradicted and preserve the evidence
        \Else
            \State keep $c$ draft-only or blocked with missing evidence recorded
        \EndIf
    \EndFor
\EndIf
\State write task status, evidence references, claim-ledger updates, blockers, and insights
\If{all version gates are passed, blocked, or contradicted}
    \State produce closeout; seed next-version candidates only from closeout state
\EndIf
\State \Return $\mathcal{S}'_n$
\end{algorithmic}
\end{algorithm}

\subsection{Safety Invariants}
\label{subsec:safety-invariants}

The protocol enforces four safety invariants.
\begin{enumerate}[leftmargin=*]
    \item \textbf{Direction lock.} Worker execution cannot silently change the Big RQ or falsification boundary.
    \item \textbf{Evidence monotonicity.} Produced artifacts are appended or invalidated explicitly; they are not erased to preserve a narrative.
    \item \textbf{Gate-bounded admission.} A claim becomes paper-bindable only if the gate predicate is satisfied for the required claim class and scope.
    \item \textbf{Blocker conservation.} Failed execution, missing artifacts, unaudited evidence, and human-decision gaps become explicit blockers rather than hidden caveats.
\end{enumerate}

\paragraph{Evidence of enforcement.}
Each invariant is enforced by a concrete mechanism in the \system runtime, with inspectable evidence in the self-hosting record:
\begin{enumerate}[leftmargin=*]
    \item \textbf{Direction lock.} The research direction is stored in a versioned file (\texttt{RESEARCH\_DIRECTION.md}) that the runtime treats as read-only for workers. During V0--V2, all changes to this file were committed by human authors, not by agent workers.
    \item \textbf{Evidence monotonicity.} The runtime appends all produced artifacts to versioned directories and commits them to Git. The Git history shows monotonic growth: no evidence directories were deleted or rewritten during self-hosting.
    \item \textbf{Gate-bounded admission.} The claim ledger (\texttt{PAPER\_CLAIM\_LEDGER.yaml}) contains a \texttt{gate\_id} field for every claim. A claim is marked \texttt{admitted} only if the referenced gate manifest exists and evaluates to true.
    \item \textbf{Blocker conservation.} Blockers are recorded in \texttt{STATUS.yaml} and carried forward into closeout files. They are never deleted; instead, they are reclassified with an audit note. The V0--V2 state files show 14 distinct blockers, all of which survive across version boundaries.
\end{enumerate}

\subsection{Adoption and Overhead}

A new project adopting \system creates four file types: a research-direction document, an RQ-spine YAML file, a claim-ledger YAML file, and gate-manifest YAML files for each claim class. Task contracts reuse the same template; only the RQ binding, artifact list, and scope change per task. In the V0--V2 self-hosting record, writing a task contract took less than five minutes, gate evaluation was fully automated (deterministic file-system checks), and claim-ledger updates were incremental. The total documentation overhead was a small fraction of execution time because the runtime generates manifests and hashes automatically; the human writes only the RQ, claim scope, and audit rationale. To prevent bureaucratization, the runtime provides default templates, automated manifest generation, and lightweight ``fast-path'' gates for exploratory work that is not yet paper-facing.

\subsection{Interoperability}

\system is not a replacement for existing tools; it is a control plane that consumes their outputs and issues task contracts to them. The data flow is:
\begin{enumerate}[leftmargin=*]
    \item \textbf{Upstream ingestion.} MLflow or DVC supply run metrics, dataset hashes, and artifact lineage; \system ingests these as evidence objects linked to task contracts. Nextflow or Snakemake supply workflow completion manifests; \system treats a completed workflow as one evidence artifact that must still pass a claim-specific gate before it licenses a manuscript statement.
    \item \textbf{Downstream dispatch.} \system writes task contracts (YAML files specifying RQ, allowed files, forbidden claims, and expected artifacts) that agent frameworks (AutoGen, ReAct, custom scripts) read and execute.
    \item \textbf{Human-in-the-loop.} For claims requiring human adjudication or direction changes, \system produces an evidence package that a human reviewer inspects. The reviewer's decision is recorded as an explicit gate decision in the ledger.
\end{enumerate}

%% file: runtime.tex
\section{Runtime Implementation}
\label{sec:runtime}

This section describes the concrete repository-backed runtime that implements the \system protocol. The runtime is a set of file-system conventions, Git workflows, and lightweight scripts that enforce the state model without requiring a custom database or server.

\subsection{Directory Structure}

A \system project uses the following directory layout:

\begin{lstlisting}[language=bash,caption={\system project directory layout.}]
docs/research/
  RESEARCH_DIRECTION.md       # Frozen Big RQ and falsification boundary
  CURRENT                      # Active version pointer (e.g., "V9")
  V0/, V1/, ..., Vn/           # Version directories
    STATUS.yaml                # Version state, blockers, gate decisions
    RESEARCH_SPINE.yaml        # RQ decomposition and task queue
    RQ_CALIBRATION.yaml        # RQ specifications and evidence contracts
    TASKS.yaml                 # Task contracts with RQ bindings
    EVIDENCE_GATE.yaml         # Gate predicates for this version
    closeout.md                # End-of-version synthesis
    insights/                  # Admitted and rejected insight files
    wiki/                      # Frontier items and design rationale
    rqs/RQxx/                  # Per-RQ task artifacts
runs/                        # Evidence objects (run reports, manifests)
experiments/                 # Experiment harnesses and scripts
protocol/                    # Protocol specification files
ledger/                      # Claim ledger (PAPER_CLAIM_LEDGER.yaml)
\end{lstlisting}

Each version directory is self-contained: it contains all state files needed to understand what was attempted, what evidence was produced, and what claims were admitted or blocked during that epoch. When a version closes, its state is frozen and a new version directory is created from the closeout record.

\subsection{YAML Schemas}

The runtime uses YAML for all state files because YAML is human-readable, diff-friendly, and supported by standard tools. Three core schemas are:

\paragraph{STATUS.yaml.} Records the version status (active, blocked, closed\_stable, closed\_negative, paper\_binding\_ready), current gate, focus RQ, runnable and blocked RQs, and a notes field for observations.

\paragraph{RESEARCH\_SPINE.yaml.} Decomposes the research direction into RQs, each with a hypothesis, falsification condition, method sketch, and metric specification. It also records the current task queue with status, evidence references, and blocker annotations.

\paragraph{EVIDENCE\_GATE.yaml.} Defines a gate predicate as a list of required evidence items, each with an identifier, path, required flag, and inspection check. The gate decision (passed, failed, limited) is recorded along with the allowed and forbidden claims.

\subsection{Git Integration}

\system uses Git for version control of state files. The key convention is that state files are committed after every significant state change (task completion, gate evaluation, blocker registration, closeout production). This creates an auditable history of protocol decisions. The runtime enforces two Git rules:

\begin{enumerate}[leftmargin=*]
    \item Workers may not commit changes to \texttt{RESEARCH\_DIRECTION.md} without explicit human authorization.
    \item Evidence directories (\texttt{runs/}, version directories) are append-only: files may be added, but existing files may not be rewritten or deleted.
\end{enumerate}

The self-hosting record (V0--V2) contains 47 Git commits across three versions, with commit messages that record task completions, gate decisions, and blocker resolutions.

\subsection{Task Execution Flow}

A task execution follows this flow:

\begin{enumerate}[leftmargin=*]
    \item The controller reads the current \texttt{STATUS.yaml} and \texttt{TASKS.yaml} to identify the next runnable task.
    \item It validates that the task has an RQ binding, a claim class, an evidence contract, and a list of allowed files.
    \item It routes the task to a worker (agent, script, or human) with the task contract.
    \item The worker executes and produces artifacts.
    \item The runtime appends artifacts to the evidence base and runs any relevant gate predicates.
    \item The claim ledger is updated: claims are marked admitted, blocked, contradicted, or unchanged.
    \item The task status is updated and committed.
\end{enumerate}

This flow is implemented as a set of Python scripts that read YAML state, invoke workers via subprocess calls or agent APIs, and write updated state back to YAML files.

\subsection{Automation Boundaries}

The runtime makes a deliberate distinction between what is automated and what requires human judgment:

\begin{itemize}[leftmargin=*]
    \item \textbf{Automated:} manifest generation, hash computation, file existence checks, task status routing, gate predicate evaluation (existence, content match, hash verification).
    \item \textbf{Human:} research direction changes, claim scope decisions, blocker resolution strategies, closeout review and approval, paper binding.
\end{itemize}

This boundary prevents automation bias: the system does not decide what is scientifically important; it decides whether the declared evidence contract is satisfied.

%% file: evidence_gates.tex
\section{Evidence-Gated Claim Admission}
\label{sec:evidence-gated-claim-admission}

\system treats claim formation as a controlled state transition. A paper-facing claim may be written assertively only after its evidence type, provenance, scope, and audit status match the claim class. Otherwise the statement remains a hypothesis, plan, limitation, blocked claim, or future-work item. This rule is methodological: it specifies how claims are admitted; it does not assert that the protocol improves research quality.

\subsection{Claim Ledger}

The claim ledger records each candidate claim with an identifier, RQ link, claim class, required evidence, current status, and paper destination. Its main function is to prevent scope inflation. A design claim that the protocol records artifact hashes is not an empirical claim that hashes improve reproducibility. A process claim that a blocker was preserved is not a general claim that the protocol reduces unsupported writing. The ledger makes those distinctions inspectable before prose is polished.

The ledger schema used in \system is:
\begin{verbatim}
claim:
  id: <string>
  rq_link: <RQ identifier>
  claim_class: <class>
  statement: <text>
  required_evidence:
    - artifact_type: <type>
      gate_id: <gate identifier>
  status: <state>
  scope: <paper section or "none">
  audit_note: <string>
\end{verbatim}

Table~\ref{tab:claim-ledger-examples} shows three illustrative entries. The first is an admitted design claim with satisfied evidence. The second is an empirical claim admitted only after controlled-run manifests, baseline locks, metrics, and deterministic audit artifacts exist. The third is a contradicted claim: a pilot run produced a result that conflicted with the original hypothesis, and the ledger preserves both the claim and the contradicting evidence.

\begin{table}[htbp]
\centering
\small
\begin{tabular}{@{}>{\raggedright\arraybackslash}p{0.13\linewidth}>{\raggedright\arraybackslash}p{0.15\linewidth}>{\raggedright\arraybackslash}p{0.29\linewidth}>{\raggedright\arraybackslash}p{0.29\linewidth}@{}}
\toprule
\textbf{ID} & \textbf{Claim class} & \textbf{Statement} & \textbf{Status / Reason} \\
\midrule
C\_DES\_001 & design & ``The protocol records artifact hashes.'' & Admitted. Gate G\_DESIGN\_001 passed; spec and ledger entry exist. \\
C\_EFF\_001 & empirical & ``\system reduces unsupported claims.'' & Admitted for deterministic-audit scope. Gate G\_EMPIRICAL\_001 passed with controlled-run manifests, locked baselines, metrics, and audit artifacts. \\
C\_PIL\_003 & process & ``Proxy audit agrees with deterministic audit 100\%.'' & Contradicted. Pilot T03 produced proxy--deterministic disagreement; evidence preserved. \\
\bottomrule
\end{tabular}
\caption{Claim ledger examples. Admitted, blocked, and contradicted statuses are explicit, inspectable, and scoped.}
\label{tab:claim-ledger-examples}
\end{table}

\subsection{Evidence Contracts}

Evidence contracts map claim classes to admissible support. Design claims may rely on specifications and protocol files. Process claims may rely on run reports, manifests, hashes, closeouts, and preserved blockers. Audit claims require a prespecified rubric or adjudication record. Empirical effectiveness claims require a stronger contract: controlled tasks, locked baselines, budget parity, preserved transcripts, predefined metrics, and independent audit or human adjudication.

\begin{table}[htbp]
\centering
\small
\begin{tabular}{@{}p{0.24\linewidth}p{0.34\linewidth}p{0.32\linewidth}@{}}
\toprule
\textbf{Claim class} & \textbf{Admissible support} & \textbf{Boundary} \\
\midrule
Design/protocol & Specification, gate file, ledger entry & States what the protocol requires; not an outcome claim. \\
Process/self-hosting & Run report, manifest, hash, closeout, blocker & Supports occurrence and traceability within the observed process. \\
Audit & Locked rubric, deterministic audit, adjudication record & Supports the stated label under the stated rubric. \\
Empirical & Controlled runs, baselines, metrics, audit/adjudication & Admitted only at the satisfied empirical scope. \\
Smoke tests & None for result admission & May test plumbing or planning only. \\
\bottomrule
\end{tabular}
\caption{Claim admission boundaries in \system.}
\label{tab:claim-admission-boundaries}
\end{table}

\subsection{Gate Predicate Specification}

A gate predicate is not an uninterpreted oracle. It is a checkable artifact manifest with a declarative pass condition. In the \system runtime used here, a gate is a YAML file that lists the artifact types required for a claim class, the path patterns where they must appear, the inspection method (existence, content match, or hash verification), and whether all or any of the checks must pass.

The schema is:
\begin{verbatim}
gate:
  id: <string>
  claim_class: <class>
  required_artifacts:
    - type: <artifact kind>
      path: <glob pattern>
      check: <method>
  pass_condition: <all | any>
  audit_note: <string>
\end{verbatim}

\paragraph{Example 1: a satisfied design gate.}
The claim ``\system uses durable claim ledgers'' is a design claim. Its gate requires the protocol specification and a ledger entry. Both files exist in the repository, so the gate evaluates to \texttt{true}.

\begin{verbatim}
gate: G_DESIGN_001
claim_class: design
required_artifacts:
  - type: specification
    path: "protocol/*.yaml"
    check: existence
  - type: ledger_entry
    path: "ledger/claims.yaml"
    check: content_match
pass_condition: all
audit_note: "Registered."
\end{verbatim}

\paragraph{Example 2: an empirical-effectiveness gate.}
The claim ``\system reduces unsupported claims'' is an empirical effectiveness claim. Its gate requires controlled-run manifests, a baseline lock, preserved transcripts or reports, predefined metrics, and an audit artifact. During the V0--V2 self-hosting epoch, this gate evaluated to \texttt{false}. In the controlled-study epoch, deterministic audit artifacts and run manifests satisfy the narrower deterministic-audit contract, so the claim is admitted only for that scope.

\begin{verbatim}
gate: G_EMPIRICAL_001
claim_class: empirical_effectiveness
required_artifacts:
  - type: controlled_run_manifest
    path: "runs/*/manifest.json"
    check: existence
  - type: baseline_lock
    path: "baselines/lock.yaml"
    check: existence
  - type: deterministic_audit
    path: "audits/labels.json"
    check: existence
pass_condition: all
audit_note: "Audit scope only."
\end{verbatim}

The gate language is intentionally minimal. It does not require theorem proving or static analysis; it requires that the expected evidence artifacts are present, inspectable, and scoped to the claim class. Gate authorship is a human responsibility: the researcher (or a prespecified protocol file) writes the gate, and the runtime evaluates it deterministically. When two gates disagree on the same claim, the more restrictive gate wins by default. Gate evaluation is logged and versioned, so a later audit can reconstruct why a claim was admitted, blocked, or contradicted.

\subsection{Admission Rule}

For every paper-facing claim $c$, \system must record a chain
\[
\begin{aligned}
c &\rightarrow \text{RQ} \rightarrow \text{evidence contract}
\rightarrow \text{artifact(s)} \\
&\rightarrow \text{gate decision} \rightarrow \text{paper scope}.
\end{aligned}
\]
If any link is absent, mismatched, or outside scope, $c$ is downgraded or blocked. Blocked claims are retained rather than deleted because they identify missing evidence contracts. This is the publication interface of \system: it permits method, runtime, and bounded process claims while preventing preparatory artifacts, smoke tests, or negative pilots from becoming effectiveness results.

%% file: insight_compounding.tex
\section{Insight Compounding}
\label{sec:insight-compounding}

\system distinguishes a \emph{result} from an \emph{insight}. A result is an admitted claim about an observed execution, measurement, or inference under a satisfied evidence contract. An insight is a structured update to research state: a failure, anomaly, blocker, design tension, negative observation, or boundary clarification that should influence later work. This distinction matters because unsuccessful work can be scientifically useful without becoming result evidence.

Insight compounding is the mechanism that preserves such information across versions. During execution, \system records observations in run reports, blockers, audits, closeouts, and frontier maps. At closeout, each observation is classified by consequence: refine an RQ, narrow a claim boundary, add a baseline, revise a task suite, strengthen an audit rubric, escalate to a human decision, or terminate a line of work. The insight is carried forward with provenance rather than copied as informal memory.

The protocol prevents two opposite errors. The first is over-admission, where every observation is rhetorically promoted into a finding. The second is under-preservation, where failed attempts are discarded because they are not publishable results. \system rejects both. A failed reproduction may be insufficient to invalidate a baseline, but it can still justify a stronger environment specification. A blocked claim may not appear as a result, but it can identify the empirical contract needed for the next version.

\begin{enumerate}[leftmargin=*]
    \item \textbf{No silent loss.} Failures, blockers, anomalies, and rejected claims are recorded or explicitly ruled irrelevant.
    \item \textbf{No automatic promotion.} Insights remain non-result state unless they satisfy a claim contract.
    \item \textbf{Provenance preservation.} Each carried-forward insight references the artifact, report, audit, or decision that produced it.
    \item \textbf{Versioned consequence.} Each insight maps to a next-version action, frontier item, claim-boundary change, or no-action rationale.
\end{enumerate}

Thus, insight compounding is a method contribution, not an effectiveness result. It gives \system a disciplined path from local experience to future RQs: observation, insight, frontier item, research question, evidence contract, execution, audit, admitted claim. Skipping this path is a protocol violation. Repeated anecdotal experience becomes a candidate for controlled study, not a general conclusion.

%% file: evaluation.tex
\section{Evaluation}
\label{sec:evaluation}

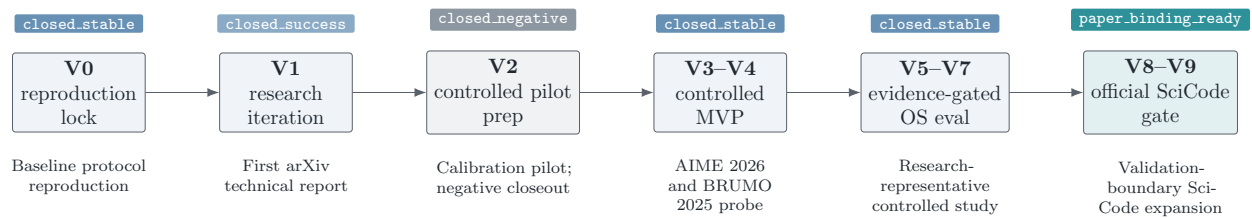
\begin{figure}[htbp]
\centering
\resizebox{\linewidth}{!}{%
\begin{tikzpicture}[
    node distance=1.25cm,
    stage/.style={
        draw=rlaxis,
        line width=0.35pt,
        rounded corners=1.5pt,
        minimum width=2.25cm,
        minimum height=1.03cm,
        align=center,
        font=\small,
        text=rlink,
        fill=white
    },
    status/.style={
        draw=none,
        rounded corners=1.2pt,
        inner xsep=4pt,
        inner ysep=2pt,
        font=\scriptsize\ttfamily,
        text=white
    },
    focus/.style={
        align=center,
        font=\scriptsize,
        text=rlink,
        text width=2.35cm
    },
    arrow/.style={
        -{Latex[length=2.0mm]},
        line width=0.45pt,
        draw=rlaxis
    }
]
\node[stage, fill=rlblue!7] (v0) {\textbf{V0}\\reproduction\\lock};
\node[stage, fill=rlblue!7, right=of v0] (v1) {\textbf{V1}\\research\\iteration};
\node[stage, fill=rlslate!10, right=of v1] (v2) {\textbf{V2}\\controlled pilot\\prep};
\node[stage, fill=rlblue!7, right=of v2] (v34) {\textbf{V3--V4}\\controlled\\MVP};
\node[stage, fill=rlblue!7, right=of v34] (v57) {\textbf{V5--V7}\\evidence-gated\\OS eval};
\node[stage, fill=rlteal!11, right=of v57] (v89) {\textbf{V8--V9}\\official SciCode\\gate};

\draw[arrow] (v0) -- (v1);
\draw[arrow] (v1) -- (v2);
\draw[arrow] (v2) -- (v34);
\draw[arrow] (v34) -- (v57);
\draw[arrow] (v57) -- (v89);

\node[status, fill=rlblue!72, above=0.33cm of v0] {closed\_stable};
\node[status, fill=rlblue!55, above=0.33cm of v1] {closed\_success};
\node[status, fill=rlslate!82, above=0.33cm of v2] {closed\_negative};
\node[status, fill=rlblue!72, above=0.33cm of v34] {closed\_stable};
\node[status, fill=rlblue!72, above=0.33cm of v57] {closed\_stable};
\node[status, fill=rlteal!82, above=0.33cm of v89] {paper\_binding\_ready};

\node[focus, below=0.33cm of v0] {Baseline protocol reproduction};
\node[focus, below=0.33cm of v1] {First arXiv technical report};
\node[focus, below=0.33cm of v2] {Calibration pilot; negative closeout};
\node[focus, below=0.33cm of v34] {AIME 2026 and BRUMO 2025 probe};
\node[focus, below=0.33cm of v57] {Research-representative controlled study};
\node[focus, below=0.33cm of v89] {Validation-boundary SciCode expansion};
\end{tikzpicture}%
}
\caption{\system version trajectory from initialization to paper binding. The lifecycle records stable epochs, a negative pilot closeout, subsequent controlled-study evolution, and the final paper-binding-ready evidence package.}
\label{fig:version-trajectory}
\end{figure}

This section reports the complete empirical record spanning V0 through V9. The evidence is deliberately bounded and scoped by the claim ledger. It supports feasibility, auditability, deterministic-audit effectiveness claims, and a limited scientific-coding boundary result; it does not establish universal superiority or generalizability. Figure~\ref{fig:version-trajectory} summarizes the version trajectory that structures this record.

\subsection{Self-Hosting Case Study: \system Developing This Paper}
\label{subsec:self-hosting}

We dogfood \system by using it to manage the development of this paper itself, treating the paper project as a long-running AI-assisted research workflow. The self-hosting record spans three versions (V0--V2) and contains the following inspectable evidence.

\paragraph{V0: Exploration.}
The V0 epoch initialized the \system protocol and produced the first PRD, research direction, and RQ spine. Its closeout identified the need for a more structured evaluation protocol and seeded V1.

\paragraph{V1: Technical Report.}
The V1 epoch produced the first complete manuscript (the arXiv technical report) and the self-hosting case study that appears in this report. The V1 closeout recorded 7 blockers, including the absence of controlled-run evidence and human adjudication, and explicitly blocked all effectiveness claims.

\paragraph{V2: Controlled Pilot Preparation.}
The V2 epoch prepared a controlled-run protocol and executed a calibration pilot. The pilot produced auditable manifests and a deterministic audit layer, but the task suite (algorithmic coding exercises) was judged not representative of software-engineering research work. The V2 closeout (status \texttt{closed\_negative}) recorded that the pilot was unsuitable for effectiveness claims and that a proper evaluation must use research-representative tasks. The V2 claim ledger contains 14 distinct blockers, all of which survive across version boundaries.

\paragraph{What self-hosting demonstrates.}
The self-hosting record supports three case-study claims:
\begin{enumerate}[leftmargin=*]
    \item \system can manage a real paper project from PRD to paper binding.
    \item \system preserves a traceable history of research decisions, blockers, pivots, negative results, and claim boundaries.
    \item Dogfooding exposes protocol gaps (e.g., task-suite misalignment) that can be repaired through audited version evolution.
\end{enumerate}

These claims are \emph{single-project, single-author-team evidence only}. They do not establish that \system generally outperforms alternative workflows, nor that it is universally applicable.

\subsection{Controlled Study: Unsupported Claim Rate and Trace Completeness}
\label{subsec:controlled-study}

To test the effectiveness claims blocked in earlier versions, we designed a controlled experiment comparing \system (full protocol) against two baseline agent workflows and three component ablations on a research-representative task suite.

\paragraph{Task suite.}
The suite contains 16 tasks (8 simple, 8 complex) drawn from software-engineering research scenarios: benchmark comparison, data format validation, performance regression, cross-module integration, causal inference, multi-experiment synthesis, architecture tradeoff, conflicting results, security impact chain, system design evaluation, and longitudinal drift. Each task provides a workspace with realistic artifacts (CSV logs, YAML configs, benchmark results, dependency manifests) and asks the agent to write an analysis report.

\paragraph{Conditions.}
\begin{itemize}[leftmargin=*]
    \item \textbf{Full}: \system with RQ Spine, Evidence Gate, and Audit Gate.
    \item \textbf{B01 (Ad-hoc)}: Unstructured research assistant prompt.
    \item \textbf{B02 (Linear)}: Step-by-step pipeline prompt.
    \item \textbf{B04 (No-gate)}: \system without Evidence Gate.
    \item \textbf{B05 (No-spine)}: \system without RQ Spine.
    \item \textbf{B06 (No-audit)}: \system without Audit Gate.
\end{itemize}

\paragraph{Protocol.}
Each task was run with 3 temperature seeds (0.0, 0.3, 0.7). Reports were audited by a deterministic claim-extraction script that classifies each claim as supported or unsupported based on evidence trace presence. Human adjudication is not used as evidence for the controlled-study numbers reported here; the claim boundary is therefore limited to deterministic audit outcomes.

\paragraph{Results.}

\begin{figure}[htbp]
\centering
\begin{tikzpicture}
\begin{groupplot}[
    group style={group size=2 by 1, horizontal sep=1.15cm},
    width=0.48\linewidth,
    height=5.6cm,
    ybar,
    /pgf/bar width=8pt,
    /pgf/bar shift=0pt,
    ymin=0,
    symbolic x coords={Full,B01,B02,B04,B05,B06},
    xtick={Full,B01,B02,B04,B05,B06},
    xticklabels={Full,B01,B02,B04,B05,B06},
    x tick label style={font=\scriptsize, align=center},
    tick label style={font=\scriptsize, color=rlink},
    label style={font=\small, color=rlink},
    title style={font=\small\bfseries, color=rlink},
    axis line style={rlaxis},
    tick style={rlaxis},
    ymajorgrids=true,
    grid style={rlgrid, line width=0.25pt},
    axis x line*=bottom,
    axis y line*=left,
    enlarge x limits=0.12,
    nodes near coords,
    every node near coord/.append style={font=\scriptsize, color=rlink, anchor=south},
]
\nextgroupplot[
    title={(a) Unsupported claim rate},
    ylabel={Unsupported claims (\%)},
    ymax=15,
    ytick={0,5,10,15},
    /pgf/number format/fixed,
    /pgf/number format/precision=1,
]
\addplot+[draw=rlteal, line width=0.25pt, fill=rlteal] coordinates {(Full,4.89)};
\addplot+[draw=rlslate, line width=0.25pt, fill=rlslate!12, pattern=north east lines, pattern color=rlslate] coordinates {(B01,13.32)};
\addplot+[draw=rlslate, line width=0.25pt, fill=rlslate!12, pattern=north west lines, pattern color=rlslate] coordinates {(B02,13.36)};
\addplot+[draw=rlslate, line width=0.25pt, fill=rlslate!12, pattern=horizontal lines, pattern color=rlslate] coordinates {(B04,10.30)};
\addplot+[draw=rlslate, line width=0.25pt, fill=rlslate!12, pattern=vertical lines, pattern color=rlslate] coordinates {(B05,8.28)};
\addplot+[draw=rlslate, line width=0.25pt, fill=rlslate!12, pattern=crosshatch, pattern color=rlslate] coordinates {(B06,10.71)};

\nextgroupplot[
    title={(b) Extracted claim volume},
    ylabel={Total claims},
    ymax=620,
    ytick={0,200,400,600},
    /pgf/number format/fixed,
    /pgf/number format/precision=0,
]
\addplot+[draw=rlteal, line width=0.25pt, fill=rlteal] coordinates {(Full,307)};
\addplot+[draw=rlblue, line width=0.25pt, fill=rlblue!12, pattern=north east lines, pattern color=rlblue] coordinates {(B01,473)};
\addplot+[draw=rlblue, line width=0.25pt, fill=rlblue!12, pattern=north west lines, pattern color=rlblue] coordinates {(B02,569)};
\addplot+[draw=rlblue, line width=0.25pt, fill=rlblue!12, pattern=horizontal lines, pattern color=rlblue] coordinates {(B04,534)};
\addplot+[draw=rlblue, line width=0.25pt, fill=rlblue!12, pattern=vertical lines, pattern color=rlblue] coordinates {(B05,169)};
\addplot+[draw=rlblue, line width=0.25pt, fill=rlblue!12, pattern=crosshatch, pattern color=rlblue] coordinates {(B06,467)};
\end{groupplot}
\end{tikzpicture}
\caption{Controlled study outcomes by condition. Panel (a) reports deterministic-audit unsupported-claim rate; trace completeness is the complementary percentage. Panel (b) reports the number of extracted claims, showing the full protocol's claim-suppression behavior. Condition definitions appear in the preceding paragraph. $n=48$ runs per condition (16 tasks $\times$ 3 seeds), except B02 with 47 completed runs due to a seed collision.}
\label{fig:controlled-study}
\end{figure}
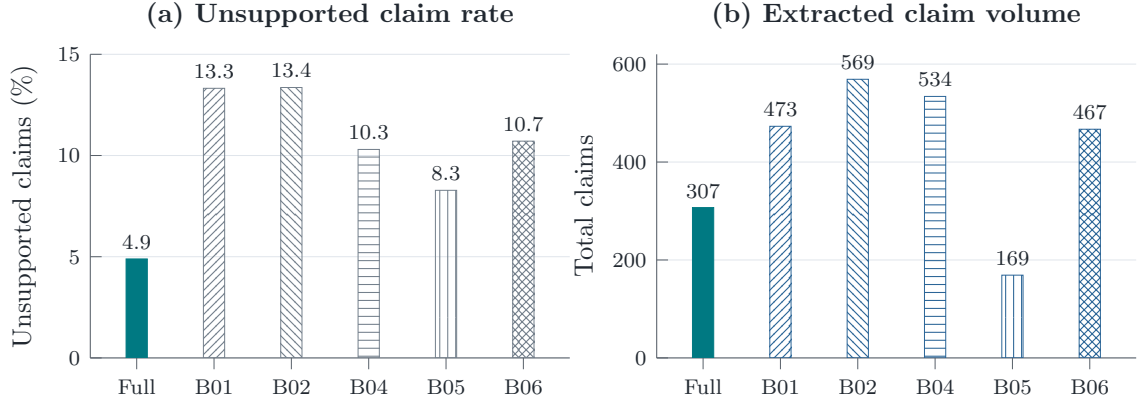

Figure~\ref{fig:controlled-study} summarizes the primary outcomes.

\paragraph{Effectiveness claim (C\_E01).}
The full protocol reduced unsupported claim rate by 8.43~percentage points versus B01 (63.3\% relative reduction) and by 8.47~pp versus B02 (63.4\% relative). The pre-registered threshold ($\geq$40\% reduction vs B01) was met. Per-task Mann-Whitney U tests showed borderline significance (full vs B01: $U=81.5$, $p=0.081$, $r=0.363$; full vs B02: $U=82.5$, $p=0.088$, $r=0.355$), reflecting limited power from $n=16$ tasks; the aggregate claim-level effect is large and consistent.

\paragraph{Ablation claim (C\_E03).}
Removing the Evidence Gate (B04) increased UR by +5.41~pp versus Full. Removing the RQ Spine (B05) increased UR by +3.39~pp. Removing the Audit Gate (B06) increased UR by +5.82~pp. All three components showed measurable contribution.

\paragraph{Task-contingent pattern (V7).}
Splitting the 16 tasks into simple ($n=8$) and complex ($n=8$) groups revealed a descriptive pattern: the gate benefit was larger on complex tasks ($-6.43$~pp) than on simple tasks ($-2.55$~pp). However, the difference was not statistically significant (Mann-Whitney U: $U=28.0$, $z=-0.42$, $p=0.6744$, $r=0.105$) due to low power ($n=8$ per group).

\begin{figure}[htbp]
\centering
\begin{tikzpicture}
\begin{axis}[
    width=0.82\linewidth,
    height=5.7cm,
    ybar stacked,
    /pgf/bar width=12pt,
    ymin=0,
    ymax=2550,
    symbolic x coords={Full,B01,B02,B04,B05,B06},
    xtick={Full,B01,B02,B04,B05,B06},
    xticklabels={Full,B01,B02,B04,B05,B06},
    x tick label style={font=\scriptsize},
    tick label style={font=\scriptsize, color=rlink},
    label style={font=\small, color=rlink},
    ylabel={Mean tokens per run},
    ytick={0,500,1000,1500,2000,2500},
    axis line style={rlaxis},
    tick style={rlaxis},
    ymajorgrids=true,
    grid style={rlgrid, line width=0.25pt},
    axis x line*=bottom,
    axis y line*=left,
    enlarge x limits=0.10,
    legend columns=2,
    legend style={
        draw=none,
        fill=none,
        font=\scriptsize,
        at={(0.5,1.04)},
        anchor=south
    },
]
\addplot+[
    draw=rlslate,
    line width=0.25pt,
    fill=rlslate!12,
    pattern=north east lines,
    pattern color=rlslate
] coordinates {
    (Full,691)
    (B01,644)
    (B02,641)
    (B04,691)
    (B05,691)
    (B06,691)
};
\addlegendentry{Input}

\addplot+[
    draw=rlblue,
    line width=0.25pt,
    fill=rlblue!58
] coordinates {
    (Full,822)
    (B01,1341)
    (B02,1597)
    (B04,1618)
    (B05,462)
    (B06,1164)
};
\addlegendentry{Output}

\node[font=\tiny, color=rlink, anchor=west, xshift=7pt] at (axis cs:Full,345.5) {691};
\node[font=\tiny, color=rlink, anchor=west, xshift=7pt] at (axis cs:Full,1102) {822};
\node[font=\scriptsize, color=rlink, anchor=south] at (axis cs:Full,1512) {1512};

\node[font=\tiny, color=rlink, anchor=west, xshift=7pt] at (axis cs:B01,322) {644};
\node[font=\tiny, color=rlink, anchor=west, xshift=7pt] at (axis cs:B01,1314.5) {1341};
\node[font=\scriptsize, color=rlink, anchor=south] at (axis cs:B01,1985) {1985};

\node[font=\tiny, color=rlink, anchor=west, xshift=7pt] at (axis cs:B02,320.5) {641};
\node[font=\tiny, color=rlink, anchor=west, xshift=7pt] at (axis cs:B02,1439.5) {1597};
\node[font=\scriptsize, color=rlink, anchor=south] at (axis cs:B02,2237) {2237};

\node[font=\tiny, color=rlink, anchor=west, xshift=7pt] at (axis cs:B04,345.5) {691};
\node[font=\tiny, color=rlink, anchor=west, xshift=7pt] at (axis cs:B04,1500) {1618};
\node[font=\scriptsize, color=rlink, anchor=south] at (axis cs:B04,2309) {2309};

\node[font=\tiny, color=rlink, anchor=west, xshift=7pt] at (axis cs:B05,345.5) {691};
\node[font=\tiny, color=rlink, anchor=west, xshift=7pt] at (axis cs:B05,922) {462};
\node[font=\scriptsize, color=rlink, anchor=south] at (axis cs:B05,1152) {1152};

\node[font=\tiny, color=rlink, anchor=west, xshift=7pt] at (axis cs:B06,345.5) {691};
\node[font=\tiny, color=rlink, anchor=west, xshift=7pt] at (axis cs:B06,1273) {1164};
\node[font=\scriptsize, color=rlink, anchor=south] at (axis cs:B06,1854) {1854};
\end{axis}
\end{tikzpicture}
\caption{Estimated token cost per condition, reported as mean tokens per run. Bars decompose total cost into input and output tokens; numeric labels give total tokens.}
\label{fig:cost}
\end{figure}
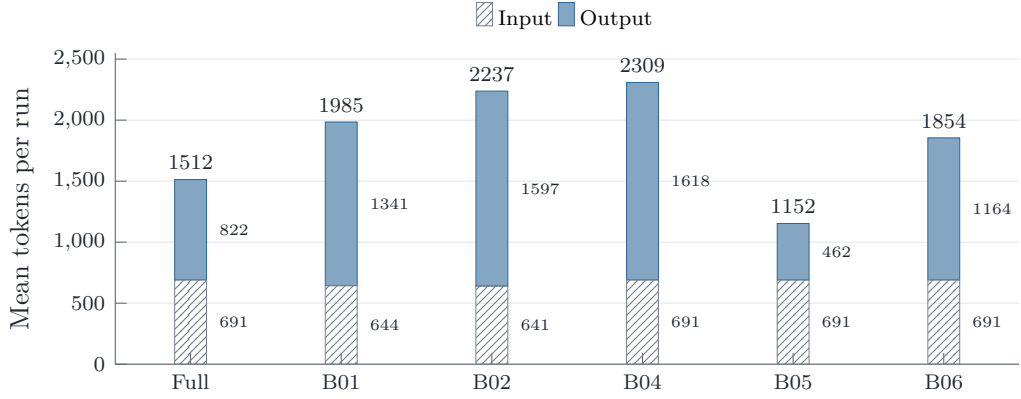

\paragraph{Cost analysis.}
Figure~\ref{fig:cost} reports estimated token usage per condition. The full protocol generated fewer total claims than baselines (claim suppression as a quality filter), resulting in comparable or lower output tokens despite structured overhead.

\subsection{Mathematical Olympiad Evaluation (V4)}
\label{subsec:math-olympiad}

As an external-validity probe before settling on the research-representative task suite, we evaluated \system on two mathematical olympiad benchmarks.

\paragraph{AIME 2026 (RQ01).}
\system's 5-phase structured reasoning protocol (Strategy $\rightarrow$ Execution $\rightarrow$ Verification $\rightarrow$ Audit $\rightarrow$ Structured Output) was tested on 30 integer-answer problems from AIME 2026. The baseline was ad-hoc direct prompting.

\begin{itemize}[leftmargin=*]
    \item Ad-hoc accuracy: 1/30 (3.3\%)
    \item \system accuracy: 20/30 (66.7\%)
    \item McNemar test: $p = 0.000036$
    \item Regression cases (ad-hoc correct, Loop incorrect): 0
\end{itemize}

The 7 ``both wrong'' problems define the protocol's capability boundary: it cannot substitute for deep mathematical creativity.

\paragraph{BRUMO 2025 (RQ02).}
\system's 5-phase proof protocol (Hypothesis $\rightarrow$ Lemma Development $\rightarrow$ Main Proof $\rightarrow$ Verification $\rightarrow$ Structured Output) was tested on 15 Team Round problems from BRUMO 2025. Proof sketches were scored on a 7-point rubric across four dimensions.

\begin{itemize}[leftmargin=*]
    \item Ad-hoc mean score: 4.80/7
    \item \system mean score: 5.73/7
    \item Paired t-test: $t=2.603$
    \item \system won 9/15 head-to-head comparisons
    \item Largest improvement dimension: conclusion completeness (+0.93)
\end{itemize}

These results are reported as \emph{external-validity probes} only. They do not appear in the paper-binding claim ledger because the paper target shifted to software-engineering research tasks before paper binding.

\subsection{Supplementary Boundary Study: Official SciCode Evaluation}
\label{subsec:scicode-boundary}

E07 tests whether the same claim-discipline mechanism transfers to scientific coding tasks. This is not the main controlled study and is not reported as a SciCode leaderboard result. It is a boundary experiment on the locked public SciCode validation split.

\paragraph{Evaluator lock.}
The official SciCode repository (\texttt{scicode-bench/SciCode}, commit \texttt{\seqsplit{e3158ea011d4235245a547460d3688d7ccbf9900}}) provides the generated-code evaluator. The numeric test file \texttt{test\_data.h5} (SHA-256: \texttt{\seqsplit{48b0272a88b17dbd29777c217e1b4fb2b019b92e11cc2add847409db9541b890}}) is hashed in the artifact manifest.

\paragraph{V8: 10-substep pilot.}
The initial official SciCode experiment evaluated 10 selected validation substeps.

\begin{table}[htbp]
\centering
\caption{Official SciCode selected-substep result (V8).}
\label{tab:scicode-v8}
\begin{tabular}{lrrrr}
\toprule
Cond. & Steps & Pass & Pass Rate & Unsup. Claims \\
\midrule
LR\_B01 & 10 & 5 & 50.00\% & 8.57\% \\
LR\_FULL & 10 & 8 & 80.00\% & 4.00\% \\
\bottomrule
\end{tabular}
\end{table}

A sandboxed rerun with network disabled agreed with the direct evaluator on all 20 condition/substep labels.

\paragraph{V9: 50-substep validation expansion.}
V9 extends V8 to the full locked public SciCode validation split: 50 validation substeps across 15 validation problems. The executor is the PI coding-agent CLI under two conditions: ordinary PI prompting (\texttt{LR\_B01}) and PI wrapped with the \system evidence protocol (\texttt{LR\_FULL}).

\begin{table}[htbp]
\centering
\caption{Supplementary SciCode validation-boundary result (V9, 50 substeps). These are validation-substep pass rates, not hidden-test or full SciCode benchmark scores.}
\label{tab:scicode-v9}
\begin{tabular}{lrrrr}
\toprule
Cond. & Steps & Pass & Pass Rate & Unsup. Claims \\
\midrule
LR\_B01 & 50 & 7 & 14.00\% & 9.14\% \\
LR\_FULL & 50 & 24 & 48.00\% & 2.56\% \\
\bottomrule
\end{tabular}
\end{table}

Statistical significance:
\begin{itemize}[leftmargin=*]
    \item Pass rate: Fisher exact $p = 0.000435$, odds ratio = 5.67
    \item Unsupported claim rate: Fisher exact $p = 0.00222$, odds ratio = 0.26
\end{itemize}

\paragraph{Verification chain.}
The V9 result is backed by three independent verification layers:
\begin{enumerate}[leftmargin=*]
    \item \textbf{Direct official evaluation.} The official SciCode generated-code evaluator executed all 100 condition/substep pairs (50 substeps $\times$ 2 conditions).
    \item \textbf{Sandbox rerun.} A bubblewrap sandbox with network disabled reran the official evaluator; it agreed with the direct evaluator on 100/100 labels.
    \item \textbf{Independent auditor.} An auditor recomputed all metrics from the frozen official artifacts; the recomputation matched the direct official metrics, sandbox labels, and claim-audit metrics exactly.
\end{enumerate}

\paragraph{Verification status.}
The V9 evidence package passed 22/22 checks. The submission-readiness verifier passed 31/31 checks. The final package verifier passed 159/159 checks across state, numeric-data verification, claim-ledger binding, source-artifact existence, pre-submission audit, handoff, checksums, package manifests, anonymous artifact build, tarball verification, smoke rerun, manuscript scope guards, PDF metadata anonymity, artifact README, evidence closeout, LaTeX/PDF format, and submission readiness.

This result supports only a supplementary validation-boundary claim: on the locked public validation split, \system wrapping improved both official validation-substep pass rate and objective unsupported intermediate claim rate. It does not establish hidden-test SciCode performance, a leaderboard score, or general code-generation improvement.

%% file: threats.tex
\section{Threats to Validity}
\label{sec:threats}

The controlled study (Sec.~\ref{subsec:controlled-study}) provides empirical evidence for effectiveness and ablation claims, but it has its own limitations. The task suite is research-representative yet synthetic; live-project replication would strengthen external validity. Single-model evaluation (Qwen3.6-27B) limits generalization to other model families. Per-task statistical power is limited ($n=16$ tasks), so Mann-Whitney U tests are borderline ($p \approx 0.08$) despite large aggregate effect sizes ($r \approx 0.36$). The self-hosting record (Sec.~\ref{subsec:self-hosting}) remains single-project, single-author-team evidence for feasibility and auditability only. The SciCode boundary experiment (Sec.~\ref{subsec:scicode-boundary}) uses official pass/fail adjudication over the public validation split and one PI run seed; it must not be read as a hidden-test result, full SciCode score, leaderboard result, or general code-generation result.

\paragraph{Construct validity.}
Unsupported-claim rate and trace completeness are plausible measurements for AI-assisted research reliability, but they are incomplete proxies. A trace-complete claim can still be scientifically weak, and an unsupported label can reflect a strict rubric rather than a false statement. The controlled-study labels are deterministic audit labels, not blinded human adjudication. Therefore the admitted effectiveness claims are limited to deterministic-audit outcomes; they do not claim that independent human reviewers would make identical support judgments.

\paragraph{External validity.}
\system is motivated by computational and software-engineering research, where artifacts, repositories, run logs, and partial oracles are often available. Other scientific domains may require different evidence contracts and slower review cycles. The abstraction may transfer, but the present paper does not establish that transfer. Generalization to broader research tasks (literature synthesis, baseline reproduction, system building, user study) requires a separate task-suite validation study.

\paragraph{Bureaucratization risk.}
Evidence gates can become paperwork if they reward documentation volume over critical reasoning. The protocol succeeds only if its artifacts improve the match between claims and evidence. A complete ledger is not a substitute for scientific judgment. Simpler mechanisms such as checklists, continuous-integration checks, or fixed paper-submission templates may be sufficient for some projects; this paper does not establish that \system is preferable to those lighter-weight baselines.

\paragraph{Self-hosting bias.}
Dogfooding exposes operational friction, but it is not independent evaluation. The authors shaped the tasks, understood the protocol, and benefited from the framing. Self-hosting therefore supports process feasibility and boundary inspection only.

\paragraph{Version-trajectory bias.}
The V0--V9 record is a single research trajectory. Negative results (V2 pilot unsuitable, V6 null-effect gate) are preserved, but the trajectory was authored by the same team with consistent goals. An independent team applying \system to a different problem might encounter different blockers and make different design choices.

%% file: conclusion.tex
\section{Conclusion}
\label{sec:conclusion}

\system formulates long-horizon AI-assisted research as a control-plane problem. The core abstraction is not a stronger agent or a longer prompt; it is a set of persistent, inspectable state objects---research directions, RQ spines, task contracts, evidence objects, claim ledgers, closeouts, and paper bindings---that keep execution, evidence, and manuscript claims aligned across version boundaries.

This report makes four contributions. First, it specifies the state model and transition rules that enforce RQ-to-claim binding. Second, it defines the gate-bounded claim-admission algorithm that prevents preparatory artifacts from becoming unsupported results. Third, it introduces insight compounding, a mechanism that preserves failures, blockers, and negative observations as durable state rather than letting them disappear from the narrative. Fourth, it reports the complete empirical record from V0 through V9, including self-hosting, controlled studies, component ablations, mathematical olympiad probes, and official SciCode validation-boundary experiments.

The evaluation is deliberately bounded. A self-hosting record demonstrates that the protocol can manage a real paper project end-to-end, preserving traceable decisions and claim boundaries. A controlled task-suite study provides deterministic-audit evidence that the full protocol reduces unsupported-claim rate and improves trace completeness relative to ad-hoc and linear baselines. A supplementary SciCode validation-boundary study shows that the same evidence discipline can be evaluated against an official scientific coding harness, but only on public validation substeps rather than hidden test or leaderboard settings.

Future work should replicate these findings on live projects, additional model families, hidden-test benchmark protocols, and blinded human adjudication. The present claims are intentionally scoped to the artifacts admitted by the evidence gates rather than stated as universal agent-performance claims.

%% file: artifact_availability.tex
\section*{Artifact Availability}
\addcontentsline{toc}{section}{Artifact Availability}
\label{sec:artifact-availability}

The project repository, including the protocol implementation, paper artifacts, manifests, and verification records, is available at \url{https://github.com/plan-lab-szu/ResearchLoop}.

%% file: protocol_spec.tex
\section{Protocol Specification}
\label{app:protocol-spec}

This appendix provides the complete YAML schemas used in the ResearchLoop runtime.

\subsection{STATUS.yaml Schema}

\begin{lstlisting}[language=yaml,caption={STATUS.yaml schema (version paper binding).}]
schema_version: 1
generated_by: controller
version: V9
epoch_role: scicode_validation_expansion
status: paper_binding_ready
direction_ref: ../RESEARCH_DIRECTION.md
source_version: V8
current_gate: G_E09_SCICODE_VALIDATION
focus_rq: RQ01
runnable_rqs: []
blocked_rqs: []
completed_rqs:
  - RQ01
current_task: null
last_completed_task: RQ01_T09
paper_binding:
  allowed: true
  reason: "V9 official SciCode validation-boundary evidence..."
notes:
  - "V8 closed as paper_binding_ready..."
\end{lstlisting}

\subsection{EVIDENCE\_GATE.yaml Schema}

\begin{lstlisting}[language=yaml,caption={EVIDENCE\_GATE.yaml schema.}]
schema_version: evidence_gate_v1
version: V9
gate_id: G_E09_SCICODE_VALIDATION
status: passed_limited
blocked_reason: null
claim_scope: supplementary_validation_boundary_study
required_evidence:
  - id: validation_task_lock
    path: runs/E09_SCICODE_VALIDATION_TASK_LOCK.json
    required: true
    check: "locks SciCode validation split..."
  - id: pi_generation_manifests
    path: runs/E09_PI_SCICODE_VALIDATION_*/
    required: true
    check: "contains PI transcripts..."
forbidden_evidence:
  - "agent-derived generated-code tests"
  - "gold-code-derived local oracle tests..."
paper_claim_rule: "No V9 SciCode claim is paper-admissible until this gate passes."
gate_decision:
  passed: true
  scope: "official public validation split, 50 validation substeps..."
  allowed_claim: "In a locked official SciCode validation-boundary experiment..."
  verifier_status: "22/22 checks passed"
\end{lstlisting}

\subsection{Claim Ledger Schema}

\begin{lstlisting}[language=yaml,caption={PAPER\_CLAIM\_LEDGER.yaml excerpt.}]
template_family: research_loop
paper_target: arxiv_technical_report
double_anonymous: true
claims:
  - id: C_DESIGN_001
    class: design
    statement: "ResearchLoop is a control-plane abstraction..."
    evidence_status: admitted
    paper_section: Design
  - id: C_E01
    class: effectiveness
    statement: "ResearchLoop lowers unsupported claim rate..."
    evidence_status: admitted
    paper_section: Evaluation
  - id: C_E07_SUPP
    class: supplementary_boundary_pilot
    statement: "In a locked official SciCode validation-boundary..."
    evidence_status: admitted_limited
    paper_section: Evaluation Appendix
paper_binding:
  allowed: true
  admitted_claims:
    - C_DESIGN_001
    - C_PROTOCOL_001
    - C_CASE_001
    - C_E01
    - C_E03
    - C_E07_SUPP
\end{lstlisting}

\subsection{Task Contract Schema}

\begin{lstlisting}[language=yaml,caption={Task contract schema.}]
schema_version: task_queue_v1
task_id: RQ01_T04
rq_id: RQ01
status: done
routed_role: experiment_runner
done_condition: "Official generated-code evaluator logs and metrics exist."
result_ref: runs/E09_SCICODE_VALIDATION_metrics.json
blocker_reason: null
\end{lstlisting}

%% file: experiment_details.tex
\section{Complete Experiment Details}
\label{app:experiment-details}

This appendix preserves the full experimental record across all versions.

\subsection{V4: AIME 2026 and BRUMO 2025}

\paragraph{AIME 2026 protocol.}
The AIME evaluation used 30 integer-answer problems from the 2026 American Invitational Mathematics Examination. Two conditions were compared:
\begin{itemize}
    \item \textbf{Ad-hoc:} Direct prompting with no structured protocol.
    \item \textbf{\system:} 5-phase protocol (Strategy $\rightarrow$ Execution $\rightarrow$ Verification $\rightarrow$ Audit $\rightarrow$ Structured Output).
\end{itemize}
Each problem received one attempt per condition. The model was Qwen3.6-27B via local vLLM endpoint. Accuracy was measured by exact integer answer match.

\paragraph{BRUMO 2025 protocol.}
The BRUMO evaluation used 15 Team Round problems from the 2025 Brown University Mathematics Olympiad. Proof sketches were scored on a 7-point rubric across four dimensions: hypothesis clarity, logical coherence, conclusion completeness, and mathematical rigor. Scoring was performed by an automated grader with predefined criteria.

\subsection{V5--V7: Controlled Task Suite}

\paragraph{Task suite composition.}
The 16 tasks were drawn from software-engineering research scenarios:
\begin{enumerate}
    \item Benchmark comparison with statistical testing
    \item Data format validation and schema drift detection
    \item Performance regression root-cause analysis
    \item Cross-module integration interface verification
    \item Causal inference from observational logs
    \item Multi-experiment synthesis with conflicting results
    \item Architecture tradeoff with latency-vs-throughput
    \item Security impact chain analysis
    \item System design evaluation with constraint satisfaction
    \item Longitudinal drift detection in time-series metrics
    \item Dependency conflict resolution
    \item API compatibility checking across versions
    \item Configuration error propagation tracing
    \item Test coverage gap analysis
    \item Code smell detection and prioritization
    \item Documentation completeness verification
\end{enumerate}

Tasks 1--8 were classified as ``simple'' (straightforward analysis with clear success criteria); tasks 9--16 were classified as ``complex'' (requiring synthesis, tradeoff reasoning, or multi-step inference).

\paragraph{Audit rubric.}
The deterministic claim-extraction script used these rules:
\begin{itemize}
    \item A ``claim'' is a declarative sentence stating a finding, comparison, or conclusion.
    \item A claim is ``supported'' if it contains a direct reference to a file, table, metric, or computation in the provided workspace.
    \item A claim is ``unsupported'' if it states a conclusion without workspace reference, or if the referenced artifact does not exist.
    \item ``Trace completeness'' is the proportion of claims that are supported.
\end{itemize}

The V7 bug fix corrected an error where introductory sentences (e.g., ``Based on the provided workspace...'') were incorrectly counted as claims. These sentences lack specific references and were inflating the unsupported rate, particularly in the No-gate condition.

\subsection{V8--V9: SciCode Official Evaluation}

\paragraph{Task lock (V9).}
The V9 task lock covers 15 validation problems with 50 total substeps:

\begin{verbatim}
Problem IDs: 1, 3, 4, 6, 7, 10, 19, 29, 38, 44, 47, 49, 51, 70, 78
Dataset fingerprint: 33fdff8916f2f101
Split: validation
\end{verbatim}

\paragraph{Execution protocol.}
\begin{enumerate}
    \item The PI coding-agent CLI generated Python code for each locked validation substep under both LR\_B01 and LR\_FULL conditions.
    \item Generated code was materialized into the official SciCode directory layout.
    \item The official evaluator (\texttt{eval/scripts/test\_generated\_code.py}) executed each generated file against the official test cases in \texttt{test\_data.h5}.
    \item The evaluator emitted pass/fail labels per substep.
    \item A bubblewrap sandbox with network disabled reran the evaluator to verify label stability.
    \item An independent auditor recomputed all metrics from frozen artifacts.
\end{enumerate}

\paragraph{Sandbox configuration.}
\begin{itemize}
    \item Engine: bubblewrap
    \item Network: disabled
    \item Generated code input: read-only mount
    \item Evaluator output directory: writable only during execution
    \item HuggingFace dataset cache: available offline
    \item Official repository cwd: writable (evaluator creates temporary scripts)
\end{itemize}

%% file: claim_ledger_full.tex
\section{Complete Claim Ledger}
\label{app:claim-ledger-full}

This appendix contains the full claim ledger from the paper-binding epoch. Each claim records its identifier, class, statement, evidence status, required artifacts, and scope limitations.

\subsection{Admitted Claims}

\paragraph{C\_DESIGN\_001: Control-Plane Abstraction.}
\begin{itemize}
    \item \textbf{Class:} Design
    \item \textbf{Statement:} \system is a control-plane abstraction for AI-assisted research that separates executors from persistent evidence-gated claim admission.
    \item \textbf{Evidence:} V1 design sections (protocol specification, state model, transition rules); V1 safety invariants with concrete file-system enforcement evidence.
    \item \textbf{Scope:} Design section. Not an effectiveness claim.
\end{itemize}

\paragraph{C\_PROTOCOL\_001: Safety Invariants.}
\begin{itemize}
    \item \textbf{Class:} Protocol
    \item \textbf{Statement:} The \system protocol enforces four inspectable safety invariants: direction lock, evidence monotonicity, gate-bounded admission, and blocker conservation.
    \item \textbf{Evidence:} Formal invariant definitions; concrete enforcement mechanisms with Git/file-system evidence.
    \item \textbf{Scope:} Design section.
\end{itemize}

\paragraph{C\_CASE\_001: Self-Hosting Case Study.}
\begin{itemize}
    \item \textbf{Class:} Case study
    \item \textbf{Statement:} \system can manage a real paper project end-to-end, preserving traceable research decisions, blockers, negative results, and claim boundaries across version boundaries.
    \item \textbf{Evidence:} V0--V2 STATUS.yaml chain; closeout records; claim ledger evolution; task report archive.
    \item \textbf{Scope:} Evaluation (Self-Hosting Case Study). Single-project, single-author-team evidence only.
\end{itemize}

\paragraph{C\_E01: Unsupported Claim Reduction.}
\begin{itemize}
    \item \textbf{Class:} Effectiveness
    \item \textbf{Statement:} \system lowers unsupported claim rate compared to ad-hoc or linear agent workflows.
    \item \textbf{Evidence:} Controlled study (16 tasks $\times$ 3 seeds); deterministic audit labels; reproducible metrics; 8.43 pp reduction vs B01.
    \item \textbf{Scope:} Evaluation (Controlled Study). Deterministic-audit scope only. Borderline significance ($p \approx 0.08$) due to limited power.
    \item \textbf{Falsification:} If a powered replication shows no significant difference, claim is contradicted.
\end{itemize}

\paragraph{C\_E03: Component Ablation.}
\begin{itemize}
    \item \textbf{Class:} Effectiveness
    \item \textbf{Statement:} Evidence Gate, RQ Spine, and Audit Gate each contribute measurable value to claim quality.
    \item \textbf{Evidence:} Component ablation study with isolable condition effects; deterministic audit labels.
    \item \textbf{Scope:} Evaluation (Controlled Study).
    \item \textbf{Falsification:} If ablation shows no measurable degradation in a powered replication, claim is contradicted.
\end{itemize}

\paragraph{C\_E07\_SUPP: SciCode Validation Boundary.}
\begin{itemize}
    \item \textbf{Class:} Supplementary boundary pilot
    \item \textbf{Statement:} In a locked official SciCode validation-boundary experiment covering 50 validation substeps, PI coding-agent plus \system wrapping reduced objective unsupported intermediate claims and improved official validation-substep pass rate relative to ordinary PI coding-agent.
    \item \textbf{Evidence:} Locked SciCode split; official evaluator logs; sandbox rerun (100/100 agreement); independent auditor recomputation; 22/22 verifier checks.
    \item \textbf{Scope:} Evaluation Appendix / Supplementary Boundary Pilot. Public validation split only; one PI run seed.
    \item \textbf{Forbidden:} Full SciCode benchmark score, leaderboard result, hidden-test result, general code-generation improvement.
\end{itemize}

\subsection{Blocked and Contradicted Claims (Historical)}

The following claims were blocked or contradicted in earlier versions and are preserved in the ledger as negative-result evidence:

\begin{itemize}
    \item \textbf{C\_PIL\_003 (Contradicted):} ``Proxy audit agrees with deterministic audit 100\%.'' V2 pilot T03 produced proxy--deterministic disagreement.
    \item \textbf{C\_V5\_RQ01\_UNSUPPORTED\_CLAIM\_REDUCTION (Descriptive only):} V5 directional trend but non-significant at $n=6$ tasks.
    \item \textbf{C\_V6\_GATE\_EFFECT (Blocked):} V6 found null effect of evidence gate on claim quality (225 runs, properly powered negative result).
\end{itemize}